\documentclass[sigconf]{acmart}
\usepackage{balance}
\AtBeginDocument{%
  }

\copyrightyear{2023}
\acmYear{2023}
\setcopyright{acmlicensed}
\acmConference[MM '23] {Proceedings of the 31st ACM International Conference on Multimedia}{October 29--November 3, 2023}{Ottawa, ON, Canada.}
\acmBooktitle{Proceedings of the 31st ACM International Conference on Multimedia (MM '23), October 29--November 3, 2023, Ottawa, ON, Canada}
\acmPrice{15.00}
\acmDOI{10.1145/3581783.3612002}
\acmISBN{979-8-4007-0108-5/23/10}

\begin{document}

\title{Video Infringement Detection via Feature Disentanglement and Mutual Information Maximization}

\author{Zhenguang Liu}
\affiliation{%
  \institution{Zhejiang University}
  \city{Hangzhou}
  \state{Zhejiang}
  \country{China}
}
\email{liuzhenguang2008@gmail.com}

\author{Xinyang Yu}
\authornote{Corresponding Authors}
\affiliation{%
  \institution{Zhejiang Gongshang University}
  \city{Hangzhou}
  \state{Zhejiang}
  \country{China}
}
\email{xyyu2022@gmail.com}

\author{Ruili Wang}
\affiliation{%
  \institution{Massey University}
  \city{Auckland}
  \country{New Zealand}
}
\email{ruili.wang@massey.ac.nz}

\author{Shuai Ye}
\affiliation{%
  \institution{Zhejiang University}
  \city{Hangzhou}
  \state{Zhejiang}
  \country{China}
}
\email{zzuyeh@163.com}

\author{Zhe Ma}
\affiliation{%
  \institution{Zhejiang University}
  \city{Hangzhou}
  \state{Zhejiang}
  \country{China}
}
\email{mz.rs@zju.edu.cn}

\author{Jianfeng Dong}
\authornotemark[1]
\affiliation{%
  \institution{Zhejiang Gongshang University}
  \city{Hangzhou}
  \state{Zhejiang}
  \country{China}
}
\email{dongjf24@gmail.com}

\author{Sifeng He}
\affiliation{%
  \institution{Ant Group}
  \city{Hangzhou}
  \state{Zhejiang}
  \country{China}
}
\email{hsf215kg@gmail.com}

\author{Feng Qian}
\affiliation{%
  \institution{Ant Group}
  \city{Hangzhou}
  \state{Zhejiang}
  \country{China}
}
\email{youzhi.qf@antgroup.com}

\author{Xiaobo Zhang}
\affiliation{%
  \institution{Ant Group}
  \city{Hangzhou}
  \state{Zhejiang}
  \country{China}
}
\email{ayou.zxb@antfin.com}

\author{Roger Zimmermann}
\affiliation{%
  \institution{National University of Singapore}
  \city{Singapore}
  \country{Singapore}
}
\email{rogerz@comp.nus.edu.sg}

\author{Lei Yang}
\affiliation{%
  \institution{Ant Group}
  \city{Hangzhou}
  \state{Zhejiang}
  \country{China}
}
\email{yl149505@antgroup.com}

\renewcommand{\shortauthors}{Zhenguang Liu et al.}

\begin{abstract}

  The self-media era provides us tremendous high quality videos. Unfortunately, frequent video copyright infringements are now seriously  damaging the interests and enthusiasm of video creators.  Identifying infringing videos is therefore a compelling task. Current state-of-the-art methods tend to simply feed high-dimensional mixed video features into deep neural networks and count on the networks to extract useful representations. Despite its simplicity, this paradigm heavily relies on the original entangled features and lacks constraints guaranteeing that useful task-relevant semantics are extracted from the features.

  In this paper, we seek to tackle the above challenges from two aspects: (1) We propose to disentangle  an original high-dimensional  feature into multiple sub-features, explicitly disentangling the feature into  exclusive lower-dimensional components. We expect the sub-features to encode non-overlapping semantics of the original feature and remove redundant information.
  (2) On top of the disentangled sub-features, we further learn an auxiliary feature to enhance the sub-features. We theoretically analyzed the mutual information between the label and the disentangled features, arriving at a loss that maximizes the extraction of task-relevant information from the original feature. 

  Extensive experiments on two large-scale benchmark datasets (\emph{i.e.}, SVD and VCSL) demonstrate that our method achieves 90.1\% TOP-100 mAP on the large-scale SVD dataset and also sets the new state-of-the-art on the  VCSL benchmark dataset. Our code and model have been released at \url{https://github.com/yyyooooo/DMI/}, hoping to contribute to the community.
\end{abstract}


\keywords{Video copyright infringements, neural network, mutual information}


\maketitle

\section{Introduction}
\label{sec:intro}

Videos and images have become the most popular information source for us. A person may spend hours watching online videos, and an appealing video on YouTube could attract millions of views per day. The sheer number of videos on multiple internet platforms (such as Facebook and TikTok) has also brought along the issue of copyright infringement. Frequent video copyright infringements now turn into a major concern for \emph{video creators}  and \emph{publishers}, who have to spend numerous resources to prevent plagiarism and seek remedies for infringed works.  As a result, infringing video retrieval has become an increasingly important task and has drawn intense  attention from the research community.

\begin{figure*}[ht]
  \centering
  \includegraphics[width=0.9\linewidth]{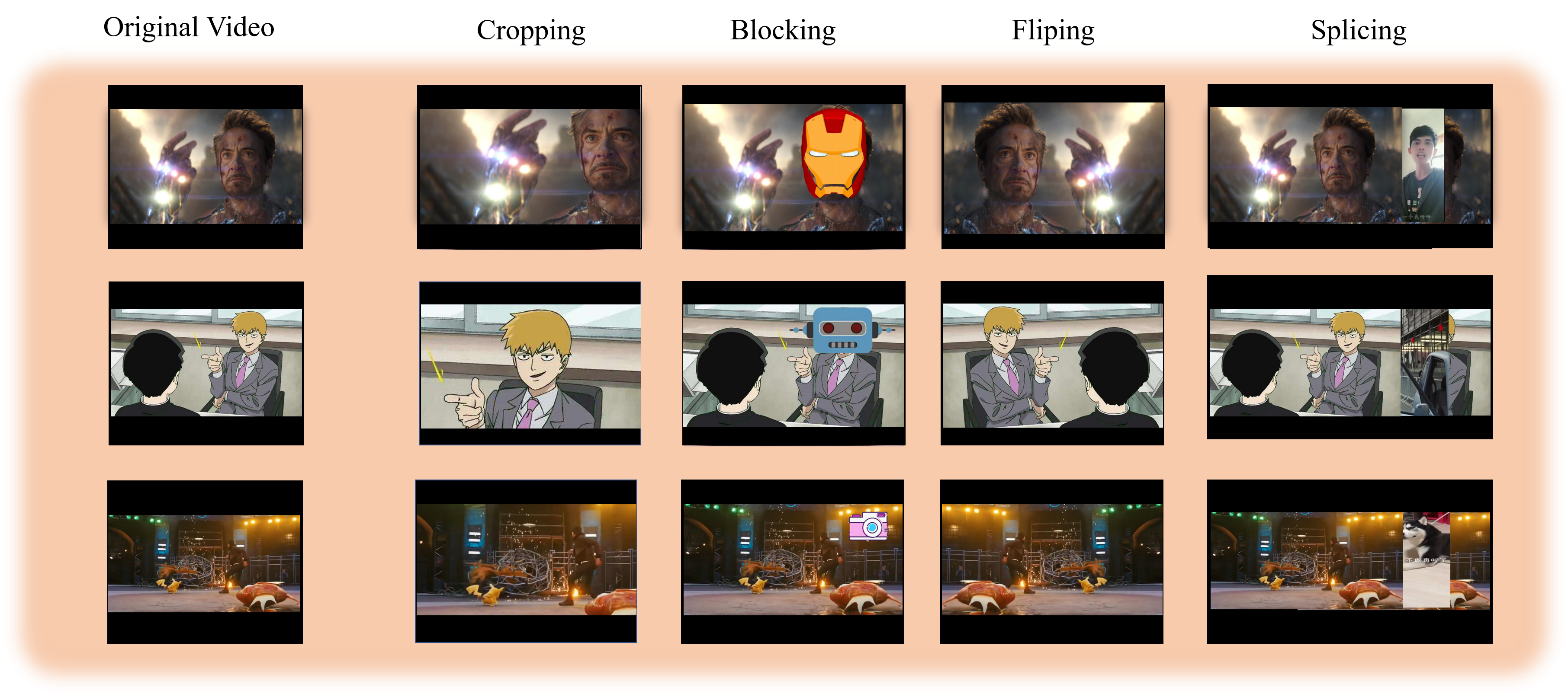}
  \caption{Video infringement usually involves video editings such as video cropping, blocking, flipping, splicing, etc. The left most column shows the original video and the right columns show the modified videos.  
}
  \label{fig:infringement_eg}
\end{figure*}

Technically, a malicious user may directly copy a video and release it on another link or website. Such duplicate videos are trivial to be detected. However, malicious users usually conduct video editings to  avoid infringement detection, the editings typically involve video cropping, blocking, flipping, splicing, \emph{etc}.  Fig.~\ref{fig:infringement_eg} demonstrates several video infringement examples. 

Infringing video retrieval amounts to detecting similar and near-duplicate video pairs out from a video gallery.  Over the past decade, \emph{code books} \cite{1999Spatial} and \emph{hashing functions} \cite{2011Multiple} have been popular solutions to generate video representations and enable video  similarity computation. However, these models heavily depend on handcrafted features and shallow functions to identify infringements, leading to unsatisfactory performance. Recently, deep learning models~\cite{yang2020tree,yang2022video,dong2018predicting,yin2021enhanced} have become the dominant approaches in infringing video retrieval, by virtue of their capacity to learn non-linear functions and sophisticated feature representations. One line of work attempts to fuse all the frame features into a single video-level representation and perform similarity matching on such representations~\cite{liu2013near, k2017near, liu2017image}. Another line of work computes a frame-to-frame similarity matrix for all the frame-level features, and subsequently combines the matrix elements into a video-level similarity score~\cite{tan2009scalable,hu2018learning,baraldi2018lamv}.

Fundamentally, given an input video pair, the essence of infringing video retrieval is to obtain their feature representations, which capture their content similarity and disregard the superficial differences ({such as \emph{varying video encodings, noises} or \emph{pixel-level differences}). Upon investigating and experimenting on the released implementations of state-of-the-art approaches \cite{SVRTN, k2017near, liu2017image}, we empirically
observe that two issues still persist: \textbf{(1)} Although existing methods have shown promising results in retrieving near-duplicate video pairs, they still have difficulties in identifying complicated cases where two videos are not explicitly similar, \emph{e.g.}, \emph{an infringing video might be obtained by cutting out part of the frame, video splicing, or blocking part of the frames}. \textbf{(2)} Current state-of-the-art methods rely on neural networks to extract label-relevant (in this task the label indicates whether two videos are similar) representations from the original feature, lacking mathematical constraints guaranteeing that  useful cues are extracted and redundant information is reduced. Consequently, the final video embedding may overfit to trivial features or insufficient representations, which compromises the performance and robustness of such methods.

In this paper, we believe it would be fruitful to investigate whether introducing supervision to disentangle the features and enforce task-relevant feature extraction would facilitate the task. With this goal in mind, we come up with the following designs. (i) Following the theoretical analysis steeped in the framework of contrastive learning and Kullback–Leibler divergence, we explicitly disentangle the original feature into a group of non-overlapping sub-features. (ii) On top of the sub-features, we further engage in mutual information theory to  extract an auxiliary feature from the original feature. The sub-features and the auxiliary feature together are composed into the final global feature. We attempt to maximize the mutual information between the label and the auxiliary feature so that no task relevant information is dropped. This framework follows a maximum relevancy and minimum redundancy paradigm, serving to remove superfluous information and retaining as much useful information as possible for the end task. 

We extensively evaluate the proposed method on two large-scale benchmark datasets (\emph{i.e.}, SVD  and VCSL). Empirical evaluations show that our approach significantly outperforms current state-of-the-art methods. Specifically, our method achieves 90.1\% TOP-100 mAP on the large SVD dataset and also sets the new state-of-the-art performance on the VCSL benchmark dataset. Our code has been submitted to Github and are availabel at \url{https://github.com/yyyooooo/DMI/} . We also present extensive ablation analyses on the contribution of each component, and evaluate the efficacy of feature disentangling and the proposed mutual information loss.

To summarize, our contributions are:

\begin{itemize}
    \item We introduce a novel framework for extracting disentangled sub-features, which incorporates KL divergence maximization and contrastive learning to supervise the sub-feature learning process. The disentangled sub-features each serves to preserve an unique aspect of the original  feature and is closely-related to the end task.
    \item In order to further extract task relevant information missed by the sub-features, we propose an auxiliary feature extraction method entrenched in the mutual information theory, which seeks to distill as much task relevant information from the original feature as possible.
    \item Extensive experiments on the large-scale  SVD and VCSL benchmark datasets show that our method surpasses state-of-the-art approaches. Interesting findings and insights on  feature disentanglement, original feature selection, and existing methods are presented.
\end{itemize}

\section{Related Work}
In this section, we  provide a brief holistic overview on the literature that is closely pertinent to this work. Roughly, the closely-related literature can be cast into 3 categories, namely \emph{video retrieval}, \emph{contrastive learning}, and \emph{mutual information theory}.

\subsection{Video Retrieval}
Earlier approaches for video retrieval mainly revolve around code books~\cite{cai2011million,kordopatis2017near,liao2018ir} and hashing functions~\cite{2011Multiple,song2013effective} for encoding  a video into a low-dimensional representation. Fueled by the success of deep learning~\cite{CVPR2021DualConsecutiveNetwork,CVPR2019MotionPrediction,DeepfakeWholebody,IJCAI2023ActionRecognition,hao2022attention,deng2023counterfactual} in recent years, the predominant approaches are to decompose the video into frames and feed them into an image extraction backbone network, obtaining a sequence of image feature representations. One approach  is to fuse all these image features into a single video-level representation and perform similar video pair detection  on video-level representations~\cite{liu2013near, k2017near, liu2017image}. Another approach is to  compute a frame-to-frame similarity matrix for all the frame-level features, and subsequently combine the matrix elements  into a video-level similarity score~\cite{tan2009scalable,hu2018learning,baraldi2018lamv}. However, current methods heavily rely on the network architecture to extract effective representations from the input video pair, with no explicit mathematical constraints that can be leveraged to enforce extraction of useful semantic cues and removal of redundancies. Our work seeks to address this research gap by proposing a new knowledge distillation framework tailored for the video retrieval task.

\subsection{Contrastive Learning}
Contrastive learning, which is typically done in a self-supervised manner, seeks to learn a representation that forces dissimilar videos to be far away from each other in the encoding space and similar videos to be close to each other \cite{he2020momentum,chen2021mocov3}. One seminal work~\cite{oord2018representation} proposes the InfoNCE loss as an effective contrastive learning objective. Furthermore, \cite{chen2020simple} presents SimCLR, which engageed in data augmentations to generate positive (similar) pairs. Another landmark work, MoCo \cite{he2020momentum}, improves upon the InfoNCE work through the use of a momentum contrast mechanism that improves convergence. \cite{DBLPISC} further advocates to  train convolutional neural networks with contrastive learning and hard data augmentation, trying to explore more discriminative representations.  Both \cite{chen2021mocov3} and  \cite{DBLPISC} employ a similar strategy of data augmentation and take an augmented image as a positive (similar) sample. Such data augmentations include \emph{cropping, grayscale, blocking part of the picture}, and \emph{horizontal flipping}, which are quite similar to the strategies adopted in video infringement. 

\subsection{Mutual Information Theory}
The mutual information of two random variables 
gives a quantitative measurement of the mutual dependency between them. 
The seminal work ~\cite{tishby2000information,tishby2015deep} of information bottleneck theory present  mathematical backgrounds of statistical learning and deep learning in the framework of mutual information.  
The theory put forth by the authors plays an indispensable role for carrying out information distillation by eliminating irrelevant input noises and preserving only those that are necessary for the task at hand.  
Estimating mutual information of two variables from their unknown distributions is notoriously challenging by nature. 
Recent works, such as~\cite{tian2021farewell,zhang2021rgb,liu2022temporal}, propose a variational self-distillation approach to estimate mutual information, which can be combined effectively with backpropagation training in the deep learning life-cycle. Inspired by this, we embrace mutual information to perform feature level supervision that constrains label-relevant representation distillation.

\section{Method}
\textbf{Problem}\quad 
Presented with a query video $q$ and a video gallery $\{g_i\}_{i=1}^n$, we are interested in retrieving all similar videos of $q$ from the video gallery. Specifically, we seek to learn an embedding for each video, where the similarity of a video pair can be conveniently computed via the distance  between their respective embedding vectors. In other words, our task can be deemed as a supervised metric learning problem where the effectiveness of the embedding is assessed based on whether similar videos have a correspondingly smaller distance in the embedding space. Unsurprisingly, the embedding vectors for dissimilar videos should be far apart and have a low similarity score.

\begin{figure*}[ht]
    \centering
    \includegraphics[width=0.9\linewidth]{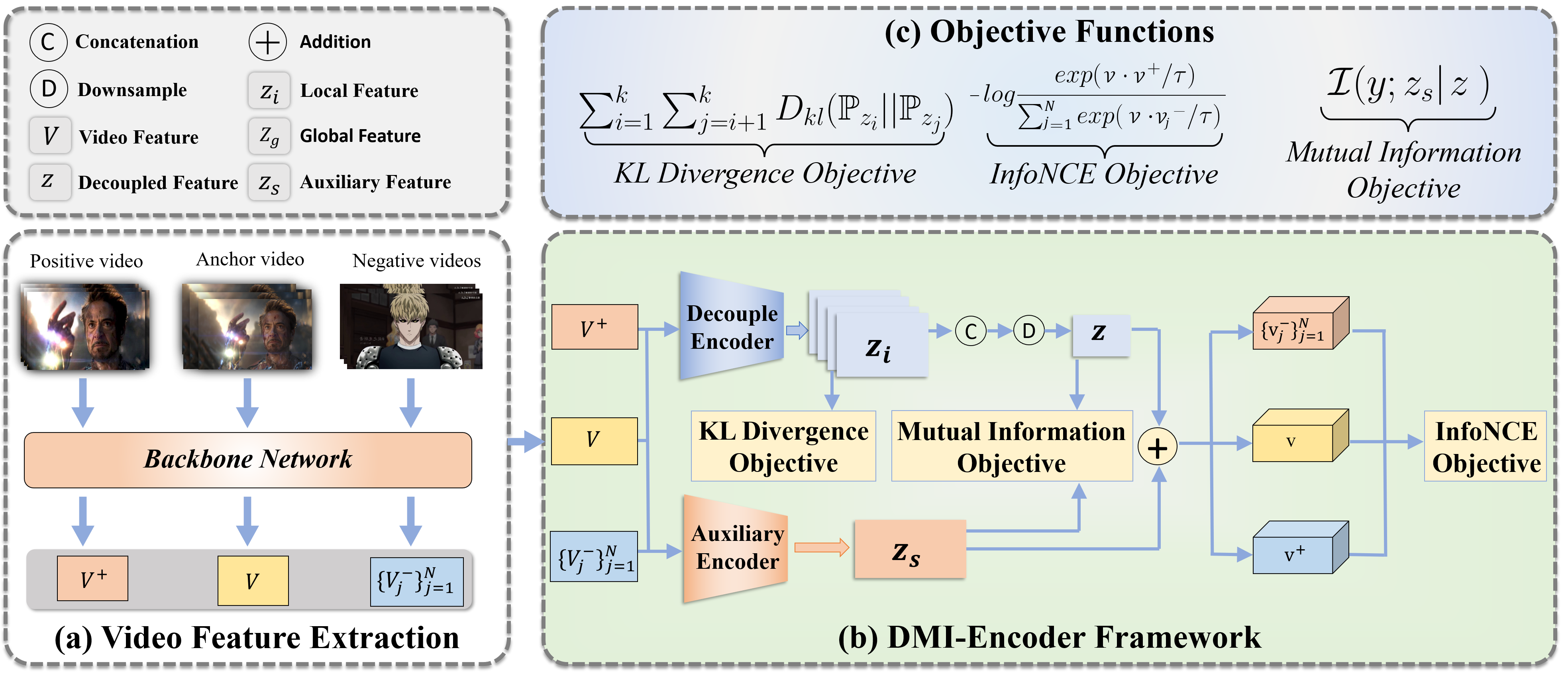}
    \caption{\textbf{Our pipeline}:  (1) In the training stage, the network input are an anchor video, a positive video (namely a video that is similar to
the anchor video), and a set of negative videos (namely a set of videos that are dissimilar to the anchor video). We utilize MoCoV3~\cite{chen2021mocov3} backbone network to extract original features for the input videos. The original features go through a disentangled feature learning module and an auxiliary  feature extraction module. The three objective functions shown in (c) constrain the network, mathematically guaranteeing the network to extract disentangled sub-features and the task-relevant auxiliary feature. (2) In the test stage, a query video is fed into the trained
network to extract its features. Videos in the gallery are also input into the newtork to obtain their features. The cosine similarity
between the query video and a video in the gallery is used to measure their similarity.}
    \label{fig:pipeline}
\end{figure*}

\subsection{Method Overview}
Broadly, we outline the proposed framework in Figure~\ref{fig:pipeline}. First, we employ a pre-trained backbone network, such as MoCoV3~\cite{he2020momentum} and VGG~\cite{simonyan2014very}, to extract a sequence of frame-level features from each input video. Subsequently, we fuse all frame-level features through average pooling, to get a video-level original feature for each video. 

Given the original feature $V$ for a video, \textbf{(1)} we propose a sub-features extraction module, which extracts a total of $k$ \textbf{disentangled} sub-features $\{\bold{z}_i\}_{i=1}^k$ from $V$. The key component in this module is to maximize the distributional distance (such as the \emph{KL divergence} and \emph{Wasserstein distance}) between each pair of sub-features. We expect each sub-feature $\bold{z}_i$ to encode exclusive aspects of the original video feature.  
As such, we break the high dimensional feature into low dimensional disentangled components to remove redundancy and facilitate the extraction of task-relevant information. \textbf{(2)} In order to ensure the sufficiency of the feature representation, we further extract an  auxiliary feature $\bold{z}_s$ to complete $\{\bold{z}_i\}_{i=1}^k$. In particular, we leverage a mutual information objective to effectuate task-relevant supplementary information extraction. The disentangled subfeatures $\{\bold{z}_i\}_{i=1}^k$ and the auxiliary feature $\bold{z}_s$ are then combined (concatenated) to arrive at our final feature representation. 
We would like to point out that the proposed approach has an edge in explicitly disentangling the feature, as well as in introducing mutual information supervision and sub-feature learning objectives.

In what follows, we will elaborate the key components of \emph{disentangled sub-feature learning} and \emph{ auxiliary feature learning}, respectively.

\subsection{Disentangled Sub-feature Learning Module}

We can certainly train a direct end-to-end CNN network for extracting infringement-related video features, as is done in most previous methods~\cite{liu2013near, k2017near, liu2017image}. However, such approaches implicitly mix different kinds of video features together, translating to high feature redundancy and low interpretability. Inspired by this, we believe it would be fruitful to investigate whether introducing supervision to disentangle the features would facilitate the task. With this goal in mind, we design a sub-features extraction module in the framework of contrastive learning and Kullback–Leibler divergence, which maps the original $d$-dimensional video feature to $k$ disentangled lower dimensional sub-features. Contrastive learning supervises  that similar videos have close embeddings and dissimilar videos have distinct embeddings, while Kullback–Leibler divergence enforces the $k$ sub-features to be different from each other.

Formally, we denote the $k$ extracted sub-features for a given anchor video $V$ as $\{\bold{z}_i\}_{i=1}^k$, and the concatenation of sub-features $\{\bold{z}_i\}_{i=1}^k$ as $\bold{v}$. For a positive video $V^+$ that is similar to $V$, we denote its concatenated sub-features as $\bold{v}^+$. Meanwhile, for a set of negative videos $\{V^-_j\}_{j=1}^N$ that are dissimilar to $V$, we denote their concatenated sub-features as $\{\bold{v}^-_1, \bold{v}^-_2, \cdots, \bold{v}^-_N\}$   respectively. 
We apply infoNCE loss \cite{oord2018representation} to $\bold{v}$, $\bold{v}^+$, and $\{\bold{v}^-_1, \bold{v}^-_2, \cdots, \bold{v}^-_N\}$, constraining that similar video pairs have similar embeddings 
while dissimilar pairs have distinct embeddings, which is given by:
\begin{equation}\label{eq_infoNCE}
    \mathcal{L}_{\text{infoNCE}}=-log\frac{exp(\bold{v}\cdot \bold{v}^+/\tau )}{{\textstyle \sum_{j=1}^{N}exp(\bold{v}\cdot \bold{v}^-_j/\tau )}},
\end{equation}
where $\tau$ is a regulation parameter.

On top of the contrastive learning, we further add supervision on the extracted sub-features $\{\bold{z}_i\}_{i=1}^k$ of a video, ensuring that different sub-features  encode minimally-overlapping aspects (semantics) of the original video. More specifically, for each two subfeatures $\bold{z}_i$ and $\bold{z}_j$ of a video, we try to maximize their difference, which is formulated as:

\begin{equation}\label{eq_decouple}
   \max~~ \mathcal{D}_\text{KL}(\mathbb{P}_{\bold{z}_{i}}||\mathbb{P}_{\bold{z}_j})= \max \Big[-\int \mathbb{P}_{\bold{z}_{i}}(x)ln(\frac{\mathbb{P}_{\bold{z}_j}(x)}{\mathbb{P}_{\bold{z}_{i}}(x)})dx\Big].
\end{equation}

Actually we have $k$ subfeatures for each video (namely $\frac{k*(k-1)}{2}$  subfeature pairs), therefore, the overall KL divergence objective is:

\begin{align}\label{eq_decouple}
  \begin{split}
    \max~~ &\sum_{i=1}^{k}\sum_{j=i+1}^k \mathcal{D}_\text{KL}(\mathbb{P}_{\bold{z}_{i}}||\mathbb{P}_{\bold{z}_j})= \\
    & \max \Big[-\sum_{i=1}^{k}\sum_{j=i+1}^k\int \mathbb{P}_{\bold{z}_{i}}(x)ln(\frac{\mathbb{P}_{\bold{z}_j}(x)}{\mathbb{P}_{\bold{z}_{i}}(x)})dx\Big].
  \end{split}
\end{align}

Through such a principled disentangling of the original feature, we are able to remove  redundancy and obtain compact disentangled sub-features, which also facilitate subsequent steps of task-relevant information extraction.

\subsection{Auxiliary Feature Extraction Module}
Up to this point, we are able to acquire the disentangled subfeatures $\{\bold{z}_i\}_{i=1}^k$ for a video. However, in the process of redundancy removing, the subfeatures might also miss some useful information in the original feature, leading to erosion of task-relevant information. Therefore, we propose to complete the subfeatures with an additional feature that consists of task-relevant information missed by subfeatures.  
To achieve this goal, we theoretically analyzed the mutual information between the label and the features, and derive a loss that maximizes the extraction of complementary task-relevant features. These allow us to enhance the extracted subfeatures and approach a better accuracy.

Formally, mutual information is a formal measure of the mutual dependency between random variables. The mutual information $\mathcal{I}(\bf{x}_{1} ; \bf{x}_{2})$ between two random variables $\bf{x}_{1}$ and $\bf{x}_{2}$ quantifies their correlated information and is defined as:
\begin{eqnarray}
\mathcal{I}(\bf{x}_{1} ; \bf{x}_{2})
= \mathbb{E}_{p(\bf{x}_{1}, \bf{x}_{2})}\left[\log \frac{p(\bf{x}_{1}, \bf{x}_{2})}{p(\bf{x}_{1}) p(\bf{x}_{2})}\right],
\end{eqnarray}
where $p(\bf{x}_{1}, \bf{x}_{2})$ is the joint probability distribution of $\bf{x}_{1}$ and $\bf{x}_{2}$,
and $p(\bf{x}_{1})$ and $p(\bf{x}_{2})$ are their marginals.

Technically, given the acquired subfeatures $\bold{z}=concatenate\,\\ \{\bold{z}_1, \bold{z}_2, \cdots, \bold{z}_k\}$, our primary objective is to maximize the amount of complementary task-relevant information in the auxiliary feature $\bold{z}_s$, which is given by: 
\begin{equation}\label{maxMI}
    \max~~\mathcal{I} (y; \bold{z}_s| \bold{z}),
\end{equation}
where $y$ denotes the label (whether two videos are similar), and $\mathcal{I} (y; \bold{z}_s| \bold{z})$ represents the amount of task-relevant information in the auxiliary feature $\bold{z}_s$.   Intuitively, maximizing this objective amounts to optimizing the additional task-relevant information we seek to extract from the original feature.

To simplify this problem, we factorize Eq.~(\ref{maxMI}) as below:

\begin{align}\label{MIfac}
    \begin{split}
    \max~~ \mathcal{I} (y ;\bold{z}_{s}| \bold{z} )\Rightarrow & \max~[ \mathcal{I} (y;\bold{z}_{s}) \\
    & -\mathcal{I} (\bold{z}_{s};\bold{z}) + \mathcal{I} (\bold{z}_{s};\bold{z}|y )].
    \end{split}
\end{align}

Interestingly, $\mathcal{I} (\bold{z}_{s};\bold{z}|y )$ represents the shared task-irrelevant information between $\bold{z}_s$ and $\bold{z}$. We may assume that the task-irrelevant information shared between $\bold{z}_s$ and $\bold{z}$ is negligible upon sufficient training \cite{tian2021farewell}. This simplifies Eq.~(\ref*{MIfac}) to:

\begin{equation}\label{sim_sd}
    \max~~  \mathcal{I} (y; \bold{z}_{s}| \bold{z})\longrightarrow \max~ [\mathcal{I} (y;\bold{z}_{s})-\mathcal{I} (\bold{z}_{s};\bold{z})].
\end{equation}

To alleviate information dropping, we further introduce a regularization term:

\begin{equation}\label{reg}
    \min\sum_{i=1}^{k} \mathcal{I}(y;\bold{z}_{i}|\bold{z}_{s}).
\end{equation}

$\mathcal{I}(y;\bold{z}_{i}|\bold{z}_{s})$ measures the vanishing task-relevant information in $\bold{z}_{i}$ during feature disentanglement.
Analogous to the procedure from Eq.~(\ref*{MIfac}) to Eq.~(\ref*{sim_sd}), we simplify the regularization term in Eq.~(\ref*{reg}) as follows:

\begin{equation}\label{final}
  \min~~  \sum_{i=1}^{k} \mathcal{I}(y;\bold{z}_{i}|\bold{z}_{s}) \longrightarrow \min~ \sum_{i=1}^{k} [\mathcal{I}(y;\bold{z}_{i}) - \mathcal{I}(\bold{z}_{i};\bold{z}_{s})].
\end{equation}

The overall mutual information objective is:
\begin{equation}\label{MI_SAMP}
    \max\mathcal{I} (y ;\bold{z}_{s}| \bold{z})+ \min\sum_{i=1}^{k} \mathcal{I}(y;\bold{z}_{i}|\bold{z}_{s}).
\end{equation}

According to Eqs.~(\ref*{sim_sd}), (\ref*{final}), and (\ref*{MI_SAMP}), the mutual information loss can be finally formulated into: 

\begin{equation}\label{uti}
    \mathcal{L}_{MI}=-[\mathcal{I} (y;\bold{z}_{s})-\mathcal{I} (\bold{z}_{s};\bold{z})] + \sum_{i=1}^{k} [\mathcal{I}(y;\bold{z}_{i}) - \mathcal{I}(\bold{z}_{i};\bold{z}_{s})].
\end{equation}

\textbf{Training Objective}. We would like to point out that the overall training objective for our end-to-end network consists of two parts.  
\textbf{(1)} Contrastive learning loss $\mathcal{L}_\text{infoNCE}$ and KL divergence loss that supervise the extraction of {disentangled sub-features}, which is given by:
\begin{align}
  \begin{split}
    \mathcal{L}_\text{decouple} =& \mathcal{L}_\text{infoNCE}  \\
    &- \alpha\cdot\Big[-\sum_{i=1}^{k}\sum_{j=i+1}^k\int \mathbb{P}_{\bold{z}_{i}}(x)ln(\frac{\mathbb{P}_{\bold{z}_{j}}(x)}{\mathbb{P}_{\bold{z}_{i}}(x)})dx\Big].
  \end{split}
\end{align}
\textbf{(2)} The mutual information loss $\mathcal{L}_{MI}$, which is presented in Eq.~(\ref{uti}), constrains the extraction of the {auxiliary feature}.
As such, the overall loss function is:
\begin{equation}
    \mathcal{L}_\text{total} = \mathcal{L}_\text{decouple} + \mathcal{L}_\text{MI}.
\end{equation}

\section{Experiments}
In this section, we evaluate the proposed approach on two large-scale benchmark datasets, SVD and VCSL. We seek to answer the
following research questions.
\begin{itemize}
\item RQ1: How is the proposed method comparing to state-of-the-art approaches?
\item  RQ2: How much do different components of our method contribute to its performance?
\item   RQ3: What interesting insights and findings can we obtain from the empirical results?
\end{itemize}
Next, we first present the experimental settings, followed by answering the research questions one by one.

\begin{figure}[ht]
    \centering
    \includegraphics[width=0.8\linewidth]{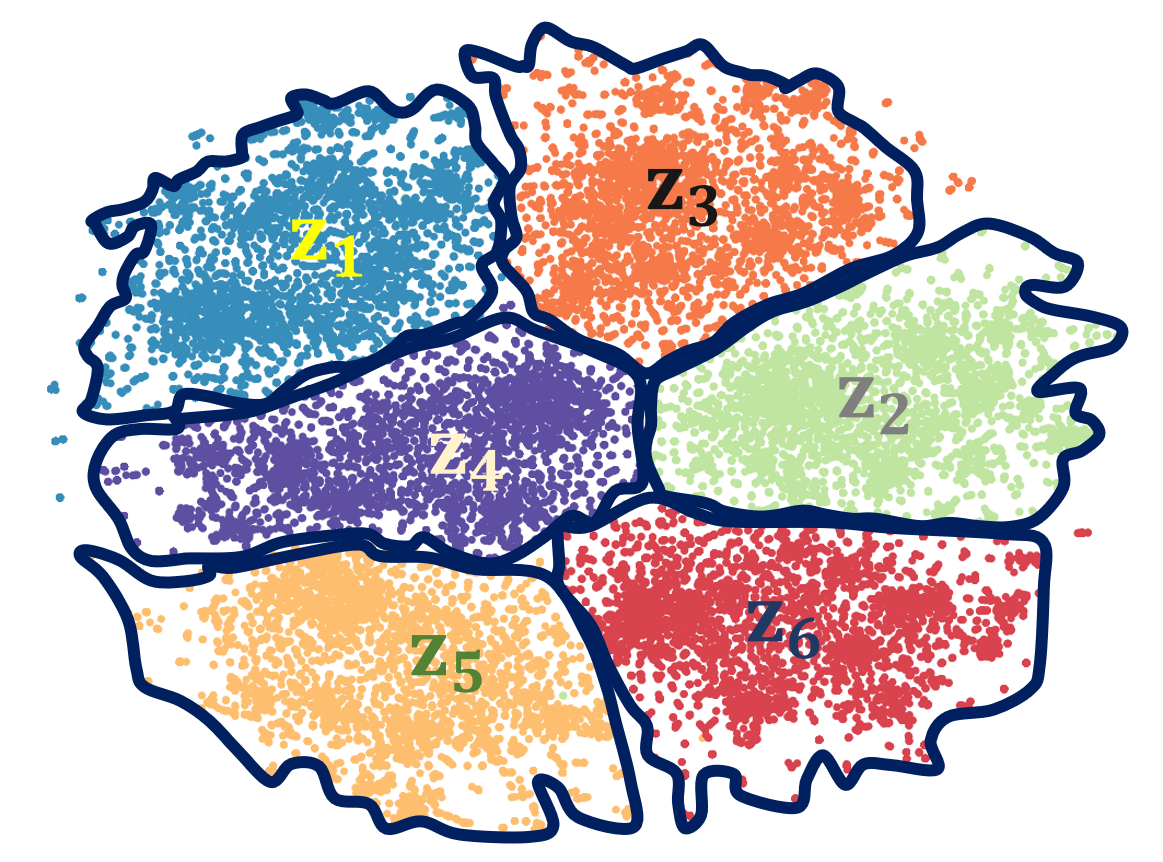}
    \caption{We visualize the features of 6,157  randomly selected videos. $\bold{z_i}$ denotes the $i^{th}$ sub-feature, which has 256 dimensions. We use TSNE to reduce the dimension of $\bold{z_i}$ to 2.}
    \label{fig:tsne}
\end{figure}

\subsection{Experimental Settings}
\textbf{Datasets}\quad The \textbf{SVD} \cite{jiang2019svd} dataset is a large-scale benchmark dataset for infringing video retrieval. The dataset contains over 500,000 videos and over 30,000 labeled pairs of infringing (similar) videos. The categories of videos are almost all-encompassing, including movies, TV series, commentary videos, daily videos, \emph{etc}.
The \textbf{VCSL } \cite{he2022large} dataset consists of more than 167,000 video infringement pairs with diverse video categories and varying video durations.

\textbf{Implementation Details}\quad Our DMI (feature \underline{D}isentanglement and \underline{M}utual \underline{I}nformation maximization) framework is implemented with PyTorch 1.4.0. For pre-processing, we divide the videos into frames and feed them into pre-trained network MoCoV3 \cite{chen2021mocov3} 
to extract the original feature for each frame. In the training process, the unlabeled videos are deemed as dissimilar to other videos and serve as negative training samples for an anchor video, which increases the number of training samples and improves the robustness of our model. Training is done with
4 Nvidia Geforce RTX 2080 Ti GPUs. The batch sizes on SVD and VCSL datasets are set to 64 and 16, respectively. In contrastive learning, the ratio of positive samples  to  negative samples for an anchor video is 64:2048. All training processes are terminated within 300 epochs. The number of sub-features is set to $k=6$. For evaluation, the widely used top-100 mAP \cite{jiang2019svd} and F-score \cite{he2022large} metrics are  adopted.

\subsection{Comparisons with State-of-the-art Methods (RQ1)}
First, we benchmark our approach against state-of-the-art methods on the \textbf{SVD dataset}. The performance of different methods are presented in terms of Top-100 mAP in Table~\ref{SVD_feature_method_cp}. A total of 8 methods are compared, including DML \cite{k2017near}, CNN-L \cite{kordopatis2017near}, CNN-V \cite{kordopatis2017near}, IsoH \cite{2012Isotropic}, ITQ \cite{gong2012ITQ}, HDML \cite{HDML}, SVRTN \cite{SVRTN}, and ours.  Specifically, {1) DML  proposes to early or late fuse frame-level features  into a single video vector, which is then fine-tuned by deep metric learning. {2) CNN-L \& CNN-V \cite{kordopatis2017near}  advocate to convert intermediate CNN features into one vector via layer and vector aggregation schemes, respectively. 3) SVRTN \cite{SVRTN} exploits a transformer structure to aggregate frame-level features into clip-level features  
 and learns the discriminative information from the interactions among clip frames. 4) ITQ \cite{gong2012ITQ} learns similarity-preserving binary codes for image retrieval by proposing an alternating minimization scheme that minimizes the quantization error.  5) IsoH \cite{2012Isotropic} improves ITQ and presents isotropic hashing strategy to learn projection functions that produce projected dimensions with isotropic variances. 6) HDML \cite{HDML}  develops a new loss-augmented inference algorithm that overcomes discontinuous optimization of the mapping from data to binary codes. 
From Table~\ref{SVD_feature_method_cp}, we observe that  metric learning-based and transformer-based approaches, namely DML and SVRTN, achieve the current state-of-the-art performance. Conventional methods \cite{gong2012ITQ,2012Isotropic,HDML},  which try to map a frame into binary codes, are surpassed by deep learning methods. Our method is  able to outperform DML and SVRTN with a 8.8\% and 3.0\% mAP gain, and overall achieves a 90.1\% TOP-100 mAP.  These  empirical evidences suggest the feasibility of the proposed approach.

\begin{table}[]
    \centering
    \caption{Comparison with existing methods on the SVD dataset.}
    \label{SVD_feature_method_cp}
  \begin{tabular}{|l|c|l|c|}
  \hline
  \textbf{Method}             & \textbf{Top-100 mAP} & \textbf{Method}             & \textbf{Top-100 mAP} \\ \hline
  \    ITQ \cite{gong2012ITQ} & 0.301                & CNN-V   \cite{kordopatis2017near} & 0.251                \\ \hline
  IsoH \cite{2012Isotropic}   & 0.309                & DML \cite{k2017near}        & 0.813                \\ \hline
  HDML \cite{HDML}            & 0.316                & SVRTN \cite{SVRTN}          & 0.871                \\ \hline
  CNN-L \cite{kordopatis2017near}   & 0.610                & \textbf{Our Method}         & \textbf{0.901}       \\ \hline
  \end{tabular}
  \end{table}

  \begin{table}[]
    \centering
    \caption{Comparison with existing methods on the VCSL dataset.}
    \label{VCSL_DETECT}
    \scalebox{1.1}{
    \begin{tabular}{|c|c|c|c|}
    \hline
    Method                                                              & F-score & Method                                                          & F-score        \\ \hline
    \begin{tabular}[c]{@{}c@{}}ViT \cite{DBLP:VIT}\end{tabular}         & 57.61   & \begin{tabular}[c]{@{}c@{}}R-MAC \cite{DBLP:rmac}\end{tabular}  & 58.75          \\ \hline
    \begin{tabular}[c]{@{}c@{}}DINO \cite{DBLP:DINO}\end{tabular}       & 59.99   & \begin{tabular}[c]{@{}c@{}}ISC \cite{DBLPISC}\end{tabular}      & 61.36          \\ \hline
    \begin{tabular}[c]{@{}c@{}}VGG \cite{simonyan2014very}\end{tabular} & 51.28   & \begin{tabular}[c]{@{}c@{}}ViSiL \cite{DBLP:ViSil}\end{tabular} & 61.46          \\ \hline
    \begin{tabular}[c]{@{}c@{}}Resnet \cite{DBLP:Resnet}\end{tabular}   & 52.07   & \begin{tabular}[c]{@{}c@{}}\textbf{Our method}\end{tabular}              & \textbf{61.69} \\ \hline
    \end{tabular}}
    \end{table}

\begin{table*}[ht]
    \begin{center}
    \begin{tabular}{|l|ll|ll|}
    \hline
    \multicolumn{1}{|c|}{\textbf{}} & \multicolumn{2}{c|}{\textbf{MoCoV3}}           & \multicolumn{2}{c|}{\textbf{VGG}}              \\ \hline
    Features                        & \multicolumn{1}{l|}{Top-100 mAP} & Top-inf mAP & \multicolumn{1}{l|}{Top-100 mAP} & Top-inf mAP \\ \hline
    Sub-feature    $\bold{z}_1$               & \multicolumn{1}{l|}{0.863}       & 0.857       & \multicolumn{1}{l|}{0.826}       & 0.805       \\ \hline
    Sub-feature $\bold{z}_2$                   & \multicolumn{1}{l|}{0.861}       & 0.855       & \multicolumn{1}{l|}{0.825}       & 0.801       \\ \hline
    Sub-feature $\bold{z}_3$                   & \multicolumn{1}{l|}{0.862}       & 0.849       & \multicolumn{1}{l|}{0.795}       & 0.778       \\ \hline
    Sub-feature $\bold{z}_4$                & \multicolumn{1}{l|}{0.867}       & 0.843       & \multicolumn{1}{l|}{0.801}       & 0.790       \\ \hline
    Sub-feature $\bold{z}_5$                  & \multicolumn{1}{l|}{0.863}       & 0.847       & \multicolumn{1}{l|}{0.824}       & 0.807       \\ \hline
    Sub-feature $\bold{z}_6$                  & \multicolumn{1}{l|}{0.851}       & 0.839       & \multicolumn{1}{l|}{0.821}       & 0.801       \\ \hline
    Global feature                  & \multicolumn{1}{l|}{\textbf{0.901}}       & \textbf{0.887 }      & \multicolumn{1}{l|}{\textbf{0.838}}       & \textbf{0.813}       \\ \hline
    \end{tabular}
    \end{center}
    \caption{Ablation study on sub-features.}
    \label{table:subfeature}
    \end{table*}

Further, we also conduct evaluations on the \textbf{VCSL dataset}. The comparison results are illustrated  in Table \ref{VCSL_DETECT}. Specifically, to evaluate the methods, we follow the F-score metric proposed by the VCSL dataset \cite{he2022large}, which serves as the official comparison metric of the dataset and measures the overlapped similar clips between two videos. Following \cite{he2022large}, we choose the Temporal Network (TN) \cite{tan2009scalable} as our alignment method. We compare our method with the approaches that achieve the current state-of-the-art performance on the VCSL dataset, including ViT \cite{DBLP:VIT}, DINO \cite{DBLP:DINO},  R-MAC \cite{DBLP:rmac}, ViSiL \cite{DBLP:ViSil}, and ISC \cite{DBLPISC}. Two baselines  VGG \cite{simonyan2014very} and Resnet \cite{DBLP:Resnet} are also included in the comparison. For VGG \cite{simonyan2014very}, we  use the VGG \cite{simonyan2014very} network to extract the  frame features and employ cosine similarity to compute the distance between frames. For Resnet \cite{DBLP:Resnet}, we use the  Resnet-50 \cite{DBLP:Resnet} network to extract frame features and adopt cosine similarity to compute the distance between frames.  
From the table, we have the following observations. Due to the inherent challenge in precisely identifying overlapped similar clips between two videos,  existing methods have yet achieved impressive F-score on this large dataset.  In particular, the current state-of-the-art methods ViSiL \cite{DBLP:ViSil} and ISC \cite{DBLPISC} obtain 61.36\%  and 61.46\%  F-score respectively on this dataset.  Our method consistently achieves the state-of-the-art performance.

\begin{table}[]
    \centering
    \caption{Ablation of different components in our method. }
    \label{SVD_ablation}\scalebox{0.7}{
\begin{tabular}{|c|c|c|c|c|c|c|}
\hline
Method                                                    & \begin{tabular}[c]{@{}c@{}}Disentangle\\ Module\end{tabular} & \begin{tabular}[c]{@{}c@{}}Auxiliary\\ Module\end{tabular} & \begin{tabular}[c]{@{}c@{}}Disentangle\\ Loss\end{tabular} & Mi Loss & \begin{tabular}[c]{@{}c@{}}Top-100\\ mAP\end{tabular} & \begin{tabular}[c]{@{}c@{}}Top-inf\\ mAP\end{tabular} \\ \hline
(a)                                                       & Remove                                                       &                                                                &                                                            &         & 0.873                                                 & 0.856                                                 \\ \hline
(b)                                                       &                                                              & Remove                                                         &                                                            &         & 0.881                                                 & 0.862                                                 \\ \hline
(c)                                                       &                                                              &                                                                & Remove                                                     &         & 0.876                                                 & 0.858                                                 \\ \hline
(d)                                                       &                                                              &                                                                &                                                            & Remove  & 0.883                                                 & 0.861                                                 \\ \hline
\begin{tabular}[c]{@{}c@{}}Default\end{tabular} &                                                              &                                                                &                                                            &         & \textbf{0.901}                                        & \textbf{0.887}                                        \\ \hline
\end{tabular}}
\end{table}

\begin{table}
    \centering
    \caption{The impact of varying the number of sub-features.}
    \label{svd_sub_layer}
    \scalebox{1.0}{
    \begin{tabular}{|c|c|c|} 
    \hline
    Feature count & top-100 mAP & top-inf mAP  \\ 
    \hline
    2           & 88.43\%     & 87.09\%      \\ 
    \hline
    4           & 88.47\%     & 86.82\%      \\ 
    \hline
    6           & \textbf{90.12}\%     &  \textbf{88.70}\%      \\ 
    \hline
    8           & 88.87\%     & 87.56\%      \\
    \hline
    \end{tabular}}
    \end{table}

\subsection{Ablation Study (RQ2)}
\textbf{Effect of the \emph{disentangled sub-feature learning} module and the \emph{auxiliary feature extraction} module.} By default, the proposed \emph{disentangled sub-feature learning} module and the \emph{auxiliary feature extraction} module are coupled together to approach high performance. We are interested in studying the impact of removing each module respectively from the proposed method. 

Towards this aim, we empirically study the impact of removing the two modules from the framework. \textbf{(a)} For the fisrt setting, we remove the sub-feature learning module in our framework, and only use the auxiliary feature extraction module to extract features. \textbf{(b)} For the second setting, we remove the global feature extraction module in our framework, and  only adopt the sub-features extraction  module to learn representations. 

\begin{figure}[!h]

  \begin{center}
  \includegraphics[width=0.8\linewidth]{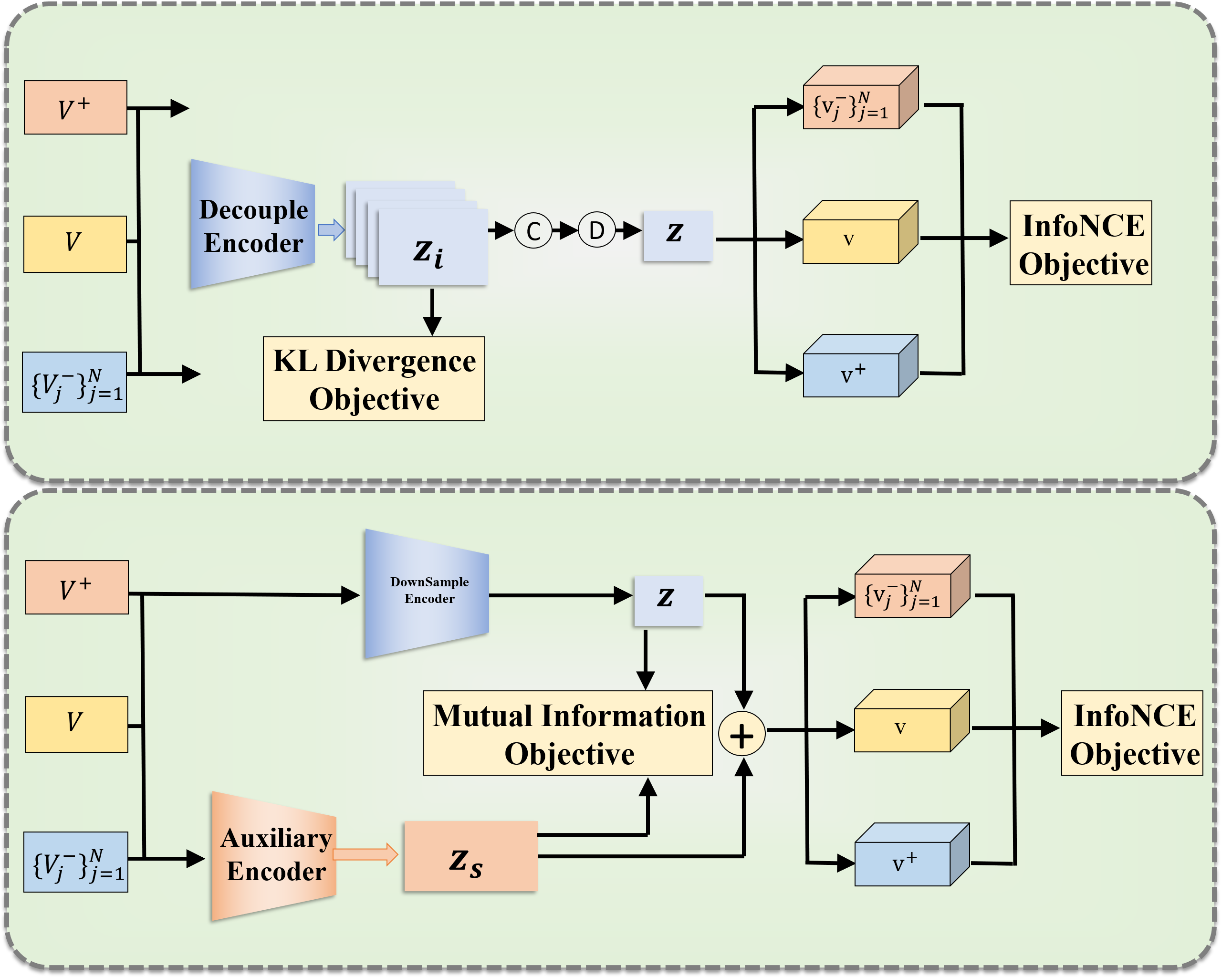}
  \end{center}
  \vspace{-1.5em}
     \caption{The architectures of the two ablation modules.}
  \label{fig:ablation}
  \vspace{-0.7em}
  \end{figure}

The two modules are shown in Fig.~\ref{fig:ablation}. When the sub-feature learning module is removed, we only compute the \emph{infoNCE} and the \emph{mutual information} objectives, where \emph{infoNCE} objective guarantees that two similar videos get similar features while the \emph{mutual information} objective enforces to extract the auxiliary feature. 

\begin{figure}[ht]
  \centering
  \includegraphics[width=1.0\linewidth]{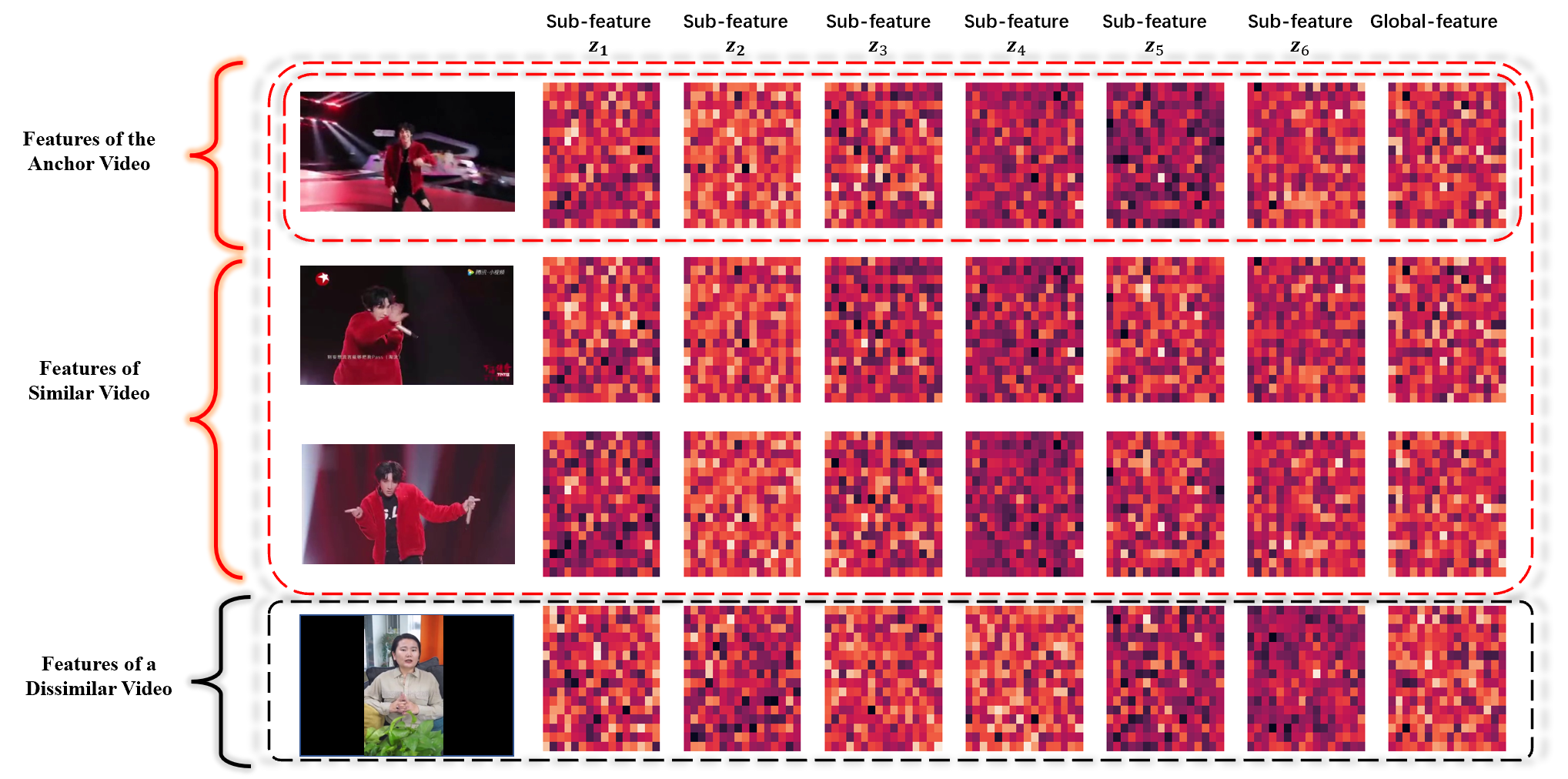}
  \caption{Features visualization for similar and dissimilar videos.}
  \label{fig:sub}
\end{figure}

We compare them with the default network. Experimental results on the SVD dataset is demonstrated in Table \ref{SVD_ablation} , where both top-100 mAP and top-inf mAP results are reported. We see that without the sub-features extraction module, the proposed method undergoes 2.8\% top-100 mAP and 3.1\% top-inf mAP drops, respectively. These empirical evidences indicate the importance of the sub-features extraction module, which contributes in removing redundancy and  disentangling features. Meanwhile, without the auxiliary feature extraction module, the method undergoes 2.0\% top-100 mAP and 2.5\% top-inf mAP drops, respectively. This suggests the significance of the auxiliary feature extraction module, which contributes in capturing more complete task-relevant information.  Removing the sub-features extraction module leads to higher performance drop than removing the auxiliary feature extraction module.

\textbf{Effect of the disengtangle loss and mutual information loss.} 
Furthermore, we study the impact of removing the proposed  disengtangle loss and mutual information loss from our network. More specifically, we remove the disengtangle loss and mutul information loss respectively from the network while preserving all other components. 
As demonstrated in \textbf{(c)} and \textbf{(d)} in Table \ref{SVD_ablation}, the top-100 mAP and top-inf mAP drop 2.5\% and 2.9\% respectively when the disengtangle loss is removed, while  top-100 mAP and top-inf mAP drop 1.8\% and 2.6\% respectively when the mutual information loss  is removed. This indicates that the disengtangle loss and mutual information loss both contribute to performance gain.

\textbf{Number of sub-features.} 
By default, in our method the number of sub-features is set to 6. It is interesting to see the effect of enlarging or reducing the number of sub-features. We illustrate the emprical results in Table \ref{svd_sub_layer}. We observe that enlarging the number of sub-laysers tends to gradualy increase the accuracy. The highest accuracy is obtained when the number of features is set to 6.

\subsection{Insights (RQ3)}
\label{sec_insights}

We make some key findings in our research. Firstly, we find that in cases of video infringement, the videos are often modified by cropping, blocking parts of the picture, splicing, and editing content. To extract the original frame features, we test the VGG and MoCoV3 (Resnet) networks as backbone networks. Our experiments showed that the MoCoV3 feature outperformed the VGG feature in terms of accuracy. We attribute this to the fact that the MoCoV3 network inherently considers data augmentations, which are commonly used in video infringement, \emph{e.g.}, cropping, grayscale, blocking parts of the picture, and horizontal flipping.

Secondly, we empirically study the separability of the extracted sub-features. To investigate this, we randomly select 6,157 videos and visualized their extracted sub-features using TSNE. The results are shown in Figure~\ref{fig:tsne}. We set $k$ to 6, so six sub-features $\{\bold{z}_i\}_{i=1}^6$ are depicted. The dimension of $\bold{z}_i$ is 256. We conduct dimension reduction using TSNE. Interestingly, we find that the sub-features are disentangled from each other.

Furthermore, we visualize the features of similar and dissimilar videos in Fig.~\ref{fig:sub}. We find that sub-features for similar videos are similar to each other, while dissimilar videos have quite different sub-features. Moreover, the sub-features for the same video are distinct from each other, which reconfirms the separability of the extracted sub-features.

We randomly sample 978 similar video pairs. We divide the 978 pairs of videos into two sets $A$ and $B$, where the two similar videos in each pair are put into two different sets. We compute the similarity matrix between the two sets of videos by utilizing the proposed method. The similarity matrix is visualized in Figure~\ref{fig:demo_gt}\textbf{(a)}, where the $x$-axis represents the 978 videos of set $A$ and $y$-axis denotes the 978 videos of set $B$. The color of a pixel $(i,j)$ reflects the similarity between the $i^{th}$ video of set $A$ and the $j^{th}$ video of set $B$. Figure~\ref{fig:demo_gt}\textbf{(b)} shows the ground-truth similarity matrix. From Figure~\ref{fig:demo_gt}\textbf{(a)} and Figure~\ref{fig:demo_gt} \textbf{(b)}, we observe that the results of our method is very close to the ground truth. Further, we zoom in the similarity matrix in Figure~\ref{fig:demo_gt}\textbf{(a)} by sampling 10 pairs of videos from the matrix. The sampled sub-matrix is demonstrated in Figure~\ref{fig:samplegt_cat_gt}\textbf{(a)}. Figure~\ref{fig:samplegt_cat_gt}\textbf{(a)} shows clearer details of the similarity matrix while  Figure~\ref{fig:samplegt_cat_gt}\textbf{(b)} provides the corresponding ground truth. We see that the results of our method are in consistent with the ground truth.

\begin{figure}[ht]
    \centering
    \includegraphics[width=0.8\linewidth]{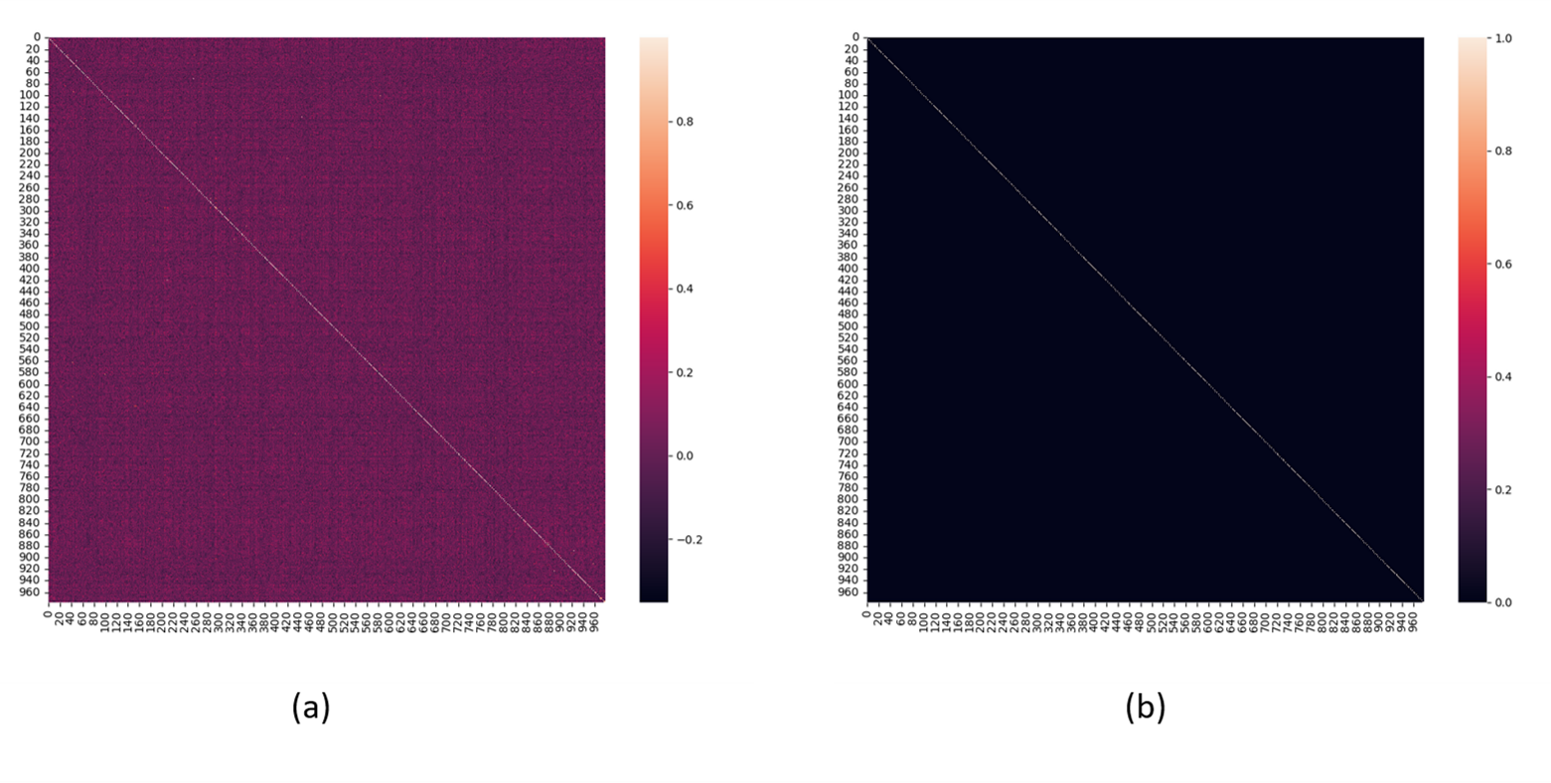}
    \caption{\textbf{(a)}: The similarity matrix computed by our method. Experiments are conducted on the SVD dataset. \textbf{(b)}: The corresponding ground-truth similarity matrix of \textbf{(a)}. }
    \label{fig:demo_gt}
\end{figure}

\begin{figure}
    \centering
    \includegraphics[width=0.75\linewidth]{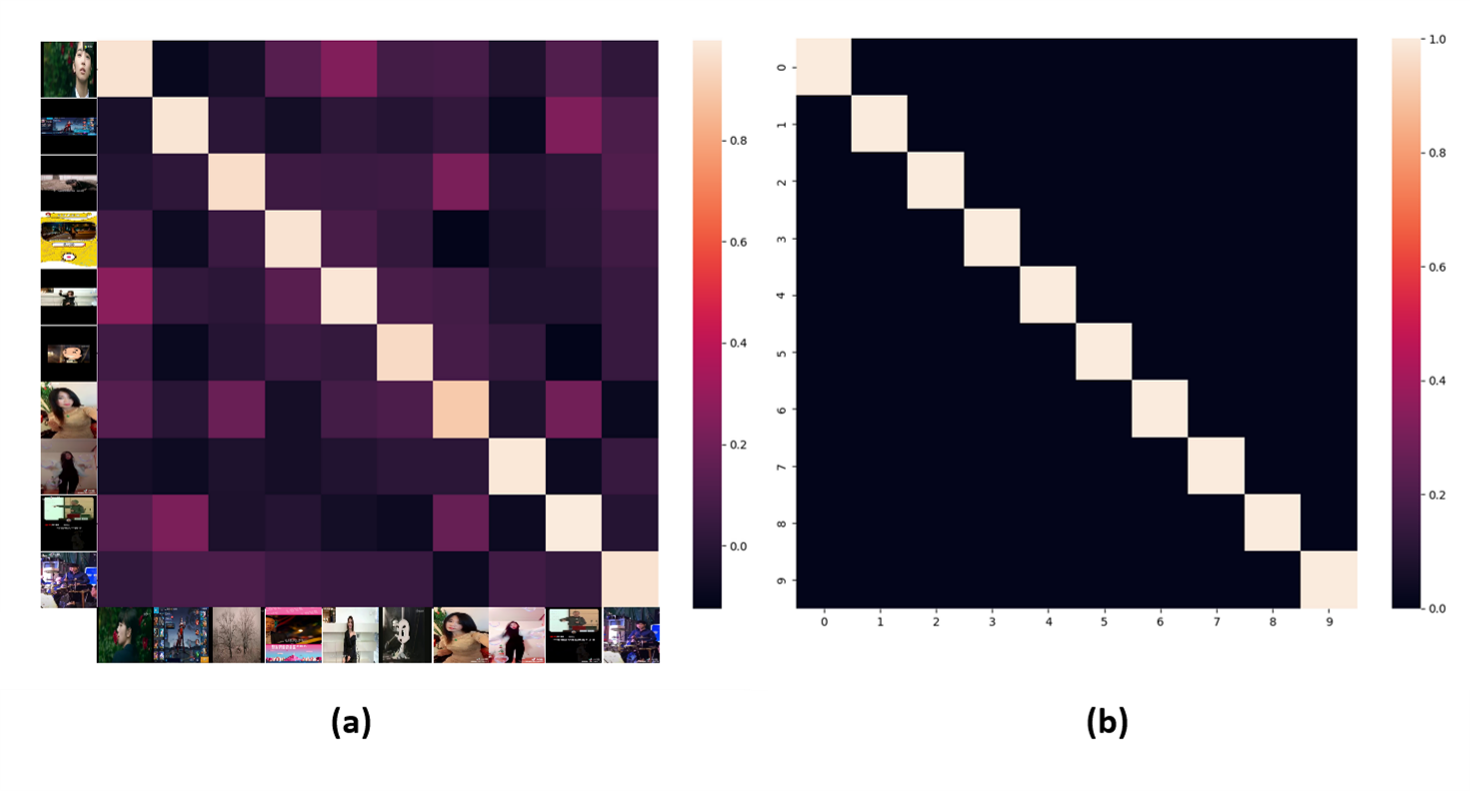}
    \caption{\textbf{(a)}: The sub-matrix sampled from the computed similarity matrix. \textbf{(b)}: The corresponding ground-truth similarity matrix of \textbf{(a)}. }
    \label{fig:samplegt_cat_gt}
\end{figure}

During the experiments, another interesting phenomena catches our attention. We are curious to know the effect of using each extracted sub-feature alone in detecting infringements. Towards this aim, we conduct experiments by  using each sub-feature alone for computing video similarity. As shown  Table \ref{table:subfeature},  we found that using only one extracted sub-feature also yields good result. However, utilizing all the sub-features and the auxiliary feature achieves better performance.

\section{Conclusion}

In this work, we have proposed a novel framework for infringing video retrieval, which is equipped with feature disentanglement and mutual information maximization. Within the framework, we combine contrastive learning and the KL divergence maximization to effectively supervise the disentangling of subfeatures. Further, we theoretically derive a mutual information objective for supervising the extraction of auxiliary and task-relevant features that might be missed by sub-features. Extensive experiments demonstrate that our method achieves state-of-the-art results on large-scale benchmark datasets, SVD and VCSL.

\begin{acks}
  This work is sponsored by Key R\&D Program of Zhejiang Province (No. 2023C01217),  CCF-AFSG Research Fund,  Zhejiang Gongshang University "Digital+" Disciplinary Construction Management Project (Project Number SZJ2022C005), and R\&D Program of DCI Technology and Application Joint Laboratory.
\end{acks}

\bibliographystyle{ACM-Reference-Format}
\balance
\bibliography{egbib}

\appendix

\end{document}